\documentclass[pmlr]{jmlr}

\usepackage{longtable}
\usepackage{enumitem}
\usepackage{booktabs}
\usepackage[load-configurations=version-1]{siunitx} 

\theorembodyfont{\upshape}
\theoremheaderfont{\scshape}
\theorempostheader{:}
\theoremsep{\newline}

\def\BibTeX{{\rm B\kern-.05em{\sc i\kern-.025em b}\kern-.08em
    T\kern-.1667em\lower.7ex\hbox{E}\kern-.125emX}}
\usepackage{csquotes}

\usepackage{hyperref}
\usepackage[capitalize,noabbrev]{cleveref}
\definecolor{mydarkblue}{rgb}{0,0.08,0.45}
\hypersetup{ %
    colorlinks=true,
    linkcolor=mydarkblue,
    citecolor=mydarkblue,
    filecolor=mydarkblue,
    urlcolor=mydarkblue,
}

\jmlrvolume{299}
\jmlryear{2025}
\jmlrworkshop{Conference on Applied Machine Learning for Information Security}

\title{Red Teaming AI Red Teaming}

\author{
\Name{Subhabrata Majumdar}
\addr {Vijil / AI Risk and Vulnerability Alliance}
\Email{subho@vijil.ai}
\\
\Name{Brian Pendleton}
\addr {AI Risk and Vulnerability Alliance}
\Email{brian@avidml.org}
\\
\Name{Abhishek Gupta}
\addr {Montreal AI Ethics Institute}
\Email{abhishek@montrealethics.ai}
}

\date{}
\editor{Edward Raff and Ethan M. Rudd}

\begin{document}
\maketitle

\begin{abstract}
Red teaming has evolved from its origins in military applications to become a widely adopted methodology in cybersecurity and AI. In this paper, we take a critical look at the practice of AI red teaming. We argue that despite its current popularity in AI governance, there exists a significant gap between red teaming's original intent as a critical thinking exercise and its narrow focus on discovering model-level flaws in the context of generative AI. Current AI red teaming efforts focus predominantly on individual model vulnerabilities while overlooking the broader sociotechnical systems and emergent behaviors that arise from complex interactions between models, users, and environments. To address this deficiency, we propose a comprehensive framework operationalizing red teaming in AI systems at two levels: macro-level system red teaming spanning the entire AI development lifecycle, and micro-level model red teaming. Drawing on cybersecurity experience and systems theory, we further propose a set of six recommendations. In these, we emphasize that effective AI red teaming requires multifunctional teams that examine emergent risks, systemic vulnerabilities, and the interplay between technical and social factors.
\end{abstract}


\section{Introduction}

In recent past, the term red teaming has gained significant attention in diverse conversations around AI as a potential solution to find and address security, safety, and reliability concerns in generative AI (genAI) systems. When ChatGPT came out in November 2022, red teaming was a relatively unknown term for AI researchers and practitioners without a background in military or cybersecurity. Less than a year later, the largest AI red teaming exercise ever was organized at DEF CON, a leading hacker conference \citep{cattell2023ai, bajak2023dont}. Fast forward to today, where we are in the middle of a (supposed) genAI revolution, AI red teaming dominates conversations in academia \citep{inie2025summon, ji2024revisiting}, think tank advice \citep{metcalf2024scaling, singh2025red}, and popular media articles \citep{burt2024how, barker2025secure, patwa2025expanding}. The notion of red teaming is being touted as a panacea to many of the problems AI might have, and its application to genAI systems has been and continues to be encouraged as a matter of practice. While that is not necessarily a bad thing---because red teaming is definitely a part of solving the AI security and safety problem---it is only a small part of a more complete set of tools that should be used.

Red teaming is a critical thinking exercise that helps determine the suitability and robustness of a proposed solution to a complex challenge. Having evolved from military war games to cybersecurity practices and now to surfacing AI security vulnerabilities and safety harms, red teaming goes beyond manual or automated testing by technical people. Rather than simply testing technological components---hardware, software, networks, and models---it extends to scrutinizing governance structures and challenging the foundational assumptions of designs from their earliest conception. As such, red teaming is not just a technical exercise, but a thought process that can be employed by nontechnical staff, such as the management, legal and risk management teams, to craft governance guidelines and engage in meaningful dialogue with technical teams about unforeseen post-deployment challenges. Sensitivity to differences in the requirements posed by safety, security, privacy, alignment, and ethics along with the application of this methodology to the total AI ecosystem of hardware, software, networks, data, and, most importantly, people will be critical for success.

There is a critical disconnect between the rich technical foundation and industry best practices in cybersecurity and the current approach to AI red teaming. Despite decades of research and education in security, today's AI red teaming efforts remain narrowly focused on technical vulnerabilities in genAI models (or model-driven general-purpose systems like ChatGPT and Claude). This missing link negates the tremendous benefits that are not realized in bridging communities of practice in AI and cybersecurity, which could ultimately lead us to finding ways to achieve robust and effective AI governance and AI solutions that adhere more closely to the espoused principles of responsible AI.

To effectively implement red teaming in a sociotechnical system like the AI development lifecycle, we need to expand our perspective beyond the narrow technical focus that currently dominates the discourse. Red teaming originated as a strategic thinking exercise in which a designated team not only simulates adversarial actions, but challenges assumptions and identifies blind spots as part of a well-defined project greenlighting process. The current approach of testing something that has already been built, as many AI companies are doing \citep{anthropic2024challenges, daws2024openai}, misses the point of pre-emptive and proactive critical analysis that red teaming was originally designed to provide.

\subsection{What Red Teaming is and is not}
Red teaming is fundamentally a team exercise \cite{zenko2015red}. It should bring together multi-functional and cross-functional teams around the common goal of ultimately improving a product or user experience. The means of this exercise are to explore hypotheticals centered around two primary questions: (1) Does the system work for their interests? (2) Could it work against their interests? The multidisciplinary approach acknowledges that systems built for a specific purpose exist within complex sociotechnical environments where technical, social, and organizational factors interact. Putting things in context, in AI red teaming technical teams are best suited to investigate model vulnerabilities. On the other hand, policy experts help identify regulatory conflicts, ethicists can surface value alignment issues, and domain specialists can evaluate real-world impact scenarios.

Although red teaming has been coopted by cybersecurity, and now AI red teams, as strictly an adversarial exercise, it should not be seen as an exercise whose goal is to capture as many clever ``gotcha" flags as possible or find ways to break the system. It is rather a collaborative process that attempts to answer the critical question: 

\begin{displayquote}
    \textit{What did we do, or not do, that could lead to failure under real-world conditions?}
\end{displayquote}

\noindent
This collaborative mindset fosters blameless transparency about blind spots, bad assumptions, and vulnerabilities. Organizations do not need to address every risk that red teamers identify, but they benefit immensely from a comprehensive awareness of potential failure modes before deployment. Effective red teaming examines not just the product itself, but the entire ecosystem: the development processes, organizational structures, incentive systems, and operational contexts in which the technology will function. In many ways, red teaming is similar to writing a research paper: the team starts with a skeleton and rough draft of the paper, then goes through cycles of redlines, experiments, and writing iterations to come up with the final version of the artifact.

\subsection{The Way Forward}

Organizations implementing AI red teaming should establish clear processes to translate findings into actionable improvements. This includes tracking identified issues, prioritizing them according to risk assessment, implementing mitigations, and verifying that the fixes do not introduce new problems. Without this feedback loop, red teaming becomes merely a performative exercise rather than a meaningful critical practice. In many ways, red teaming is similar to acceptance testing in software development. And we all know that no amount of pre-deployment testing can anticipate all potential issues. Red-teamed systems operate in dynamic environments with evolving threats and user behaviors. This reality necessitates ongoing monitoring, continuous evaluation, and adaptive response capabilities that extend well beyond the initial red teaming exercises.

When defining AI governance and risk management practices, organizations should remember that the goals of AI red teaming are broader than just ensuring secure and safe behavior of AI models, and its means are deeper than narrow technical approaches like pentesting or fuzzing. In the rest of this paper, we argue for and design red teaming as a collective critical thinking methodology that, when properly applied throughout the development lifecycle, can substantially maximize the utility and minimize the risks of an AI system. 

\paragraph{}
The remainder of this paper proceeds as follows. Section~\ref{sec:history} traces the evolution of red teaming from military war games through cybersecurity practices to current AI applications, highlighting the progressive narrowing of scope. Section~\ref{sec:implement} presents our two-level framework, distinguishing macro-level system red teaming from micro-level model testing, and provides concrete examples of each. Section~\ref{sec:system} presents a few recommendations to consider AI red teaming from a systems perspective, and Section~\ref{sec:conclusion} concludes the paper.

\section{History of Red Teaming}
\label{sec:history}

We set the context for this paper by tracing the evolution of red teaming across different fields, with a particular focus on how it has been adapted to AI systems and the limitations of red teaming in its most recent incarnation.

\subsection{Military Origins}

Red teaming emerged from the tactical war games of the Prussian military in the early 19th century, evolving through Cold War simulations into today's formalized methodology for challenging conventional wisdom and identifying strategic and exploitable hazards. The Prussian army adopted \textit{Kriegsspiel} (literally ``wargame'' in German) in 1812, a tabletop simulation developed by Lieutenant Georg Leopold von Reisswitz and his son, where blue pieces represented the Prussian forces and red represented the enemy \citep{meyer2025tenth, kay2020kriegsspiel}. This color-coding established the concept of ``red team" that persists to this day.

In the United States (US), the practice of red teaming took shape during the Cold War, when the RAND Corporation conducted military simulations for the US government. In these exercises, ``red team'' and the color red were used to represent the Soviet Union, while ``blue team'' and blue represented the United States \citep{wikipedia2025red}. Following the intelligence failures that led to the September 11 attacks in 2001, the US Department of Defense established formal red team units to prevent similar catastrophic oversights in future. The 9/11 Commission identified a ``failure to connect the dots'' as a primary cause of the intelligence breakdown. This led to systematic changes to prevent groupthink and foster alternative analysis \citep{redteamthinking2021}. As a result, the Pentagon created specialized training centers such as the University of Foreign Military and Cultural Studies at Fort Leavenworth to institutionalize red teaming methodologies. The US Army's University of Foreign Military and Cultural Studies at Fort Leavenworth, created after intelligence failures in Iraq, developed the modern framework for red teaming that transformed an ad hoc practice into a systematic methodology for critical analysis \citep{dooley2017psychology}. This program teaches military officers and government officials techniques to challenge assumptions, consider alternative perspectives, and introduce contrarian thinking into planning processes.

The concept of the ``10th man'' from Israeli military doctrine, popularized in the film ``World War Z'', illustrates another approach to institutionalized contrarian thinking. When everyone agrees on a particular outcome, it is the designated 10th man's responsibility to disagree and explore alternative scenarios. This concept reportedly developed after intelligence failures during the 1973 Yom Kippur War, when analysts unanimously agreed that Arab troop movements weren't a threat \citep{meyer2025tenth}. In reality, while Israeli intelligence did establish a unit called \textit{Ipcha Mistabra} (``the opposite side'') to challenge prevailing assumptions after the Yom Kippur War, the specific ``10th man'' concept as portrayed in the film is somewhat fictionalized but based on real adversarial thinking practices in military intelligence \citep{wikipedia2025red}.

\subsection{Adoption in Cybersecurity}

The National Security Agency (NSA) first recognized the need for proactive cybersecurity measures in the 1980s, and pioneered the concept of ``red teams'' tasked with assessing the security of classified systems \citep{techround2023history}. These early efforts involved independent evaluators simulating potential attackers and identifying weaknesses that required remediation.

As digital threats evolved in the 1990s, so did cybersecurity red teaming. The term ``tiger team'' was initially used to describe specialized groups that performed many of the same functions as modern red teams \citep{firch2024red}. These groups of expert professionals were hired to take on particular challenges against the security posture of organizations.

Following the 9/11 attacks, cybersecurity red teaming gained significant momentum as organizations recognized the need for more comprehensive security testing. The Central Intelligence Agency created a new unit called ``Red Cell'', and red teaming became increasingly common in various government agencies to model responses to asymmetric threats, including cyber attacks \citep{wikipedia2025red}. This period marked the transition from isolated pentesting to more holistic security assessments that incorporated physical security, social engineering, and other non-technical aspects.

Modern cybersecurity red teaming encompasses several key methodologies working in concert: technical assessments that test digital defenses through vulnerability scanning, exploitation, and lateral movement \citep{anderson2025what}; physical security testing that evaluates access controls for facilities \citep{sapphire2024red}; social engineering that targets the human element through phishing and impersonation \citep{tomkiel2024penetration}; and extended red team operations designed to achieve specific objectives while testing detection and response capabilities \citep{redscan2024red}. These approaches are codified in frameworks such as NIST Special Publication 800-53, which includes specific controls for red team exercises designed to ``simulate attempts by adversaries to compromise organizational information systems'' and ``provide comprehensive assessments that reflect real-world conditions'' \citep{csf2021ca}.

The field continues to evolve with several advanced approaches. Continuous Automated Red Teaming (CART) uses automation to assess security posture in real-time rather than through periodic manual assessments \citep{anderson2025what}. Adversary Emulation models tactics after specific threat actors that might target the organization, guided by frameworks like MITRE ATT\&CK \citep{cisa2024enhancing}. Purple Teaming fosters collaboration between red and blue teams to identify vulnerabilities and improve response strategies \citep{depalma2023penetration}. Integrated IT-OT Assessments expand scope to include industrial control systems and critical infrastructure \citep{frumento2021integrated}. AI-Enhanced Red Teaming that incorporates AI to improve the effectiveness of assessments \citep{kirvan2024red}. And finally, specialized services from organizations like CISA provide comprehensive evaluations for critical infrastructure sectors and government agencies \citep{cisa2024enhancing}.

This evolution of cybersecurity red teaming from isolated technical assessments to comprehensive, intelligence-driven simulations reflects the increasing sophistication of cyber threats and the growing recognition that effective security requires a holistic approach that addresses technical, physical, and human vulnerabilities.

\subsection{Red Teaming AI}

A working definition of the very new concept of AI red teaming is that it is structured testing to identify flaws and vulnerabilities in AI systems, typically conducted in controlled environments with developer collaboration. For Large Language Models (LLMs) specifically, red teaming is defined as ``a process where participants interact with the LLM under test to help uncover incorrect or harmful behaviors'' \citep{romero2025red}. LLM development companies have implemented various approaches to red teaming, ranging from comprehensive security assessments to narrower evaluations focused on specific genAI features \citep{bullwinkel2025lessons}. In the last two years, such approaches have become recognized as essential for assessing the risks of AI models and systems \citep{ahmad2025openai}.

Despite its growing popularity, researchers have identified significant challenges with current AI red teaming practices. The field lacks consensus on the scope, structure, and assessment criteria for AI red teaming \citep{feffer2024red}, raising concerns that red teaming may sometimes function more as performative activity than substantive risk mitigation. Many current approaches focus too narrowly on the models themselves, neglecting how vulnerabilities might manifest in production systems where AI models are a part of broader systems \citep{bullwinkel2025lessons}. Additionally, most AI experts fail to consider insider risks \citep{martin2025sorry}, and most testing processes remain limited to English-language evaluations \citep{romero2025red}.

The process of red teaming itself presents challenges for participants. They can experience negative psychological impacts from interaction with harmful content and adversarial thinking, which can lead to decreased productivity or psychological harm \citep{ahmad2025openai}. Many red teamers also lack training in crucial disciplines outside of their technical expertise, with most employee red teamers having limited proficiency in linguistic, sociocultural, historical, legal, or ethical domains \citep{gillespie2025ai}.

As the field matures, researchers increasingly recognize that red teaming alone cannot solve all the challenges in AI risk assessment \citep{ahmad2025openai}. More effective approaches such as violet teaming are emerging that integrate complementary methods, combining red teaming's risk identification capabilities with the solution-focused responses of blue teaming \citep{titus2023promise}. Future success will likely depend on embracing greater diversity and focusing more on practical red teaming efforts that address real-world attacks, which are often less sophisticated than those presented in academic papers \citep{rawat2024attack}.

\subsection{What's Missing?}

As such, the state-of-the-art of AI red teaming over-indexes on model-specific behaviors rather than how these models work within AI applications, interact with broader social contexts, and affect the external world through their outputs. Current AI red teaming efforts tend to focus narrowly on technical vulnerabilities while overlooking broader sociotechnical considerations. The public interest dimension of red teaming remains underdeveloped as well \citep{singh2025red}.

To mitigate these flaws, AI red teams need to embrace cybersecurity's decades of experience with security testing and reporting. A large proportion of the testing done in the name of red teaming AI systems can be structured and automated through frameworks like MITRE ATT\&CK for understanding adversarial behaviors, prioritizing vulnerabilities and coordinating defensive responses \citep{mitre2023attack}, and established practices for continuous monitoring, automated testing, and incident response \citep{nist2020security}. In comparison, effective red teaming benefits from ``an alchemist mindset" that extends beyond purely technical approaches \citep{inie2025summon}. Successful cyber red team engagements typically involve creating diverse and highly realistic scenarios that produce actionable insights to aid proactive mitigation \citep{yulianto2025enhancing}---principles that are often missing from AI red teaming.

\section{Implementation}
\label{sec:implement}

Any red team activity should be part of a larger, coordinated risk and security effort. This includes pre-mortems conducted before the development of a model and associated systems begins, and comprehensive risk assessments of the model and associated systems. Organizations should incorporate AI-specific security processes in the model development lifecycle and include relevant security teams that can harden both the model and associated systems. Finally, a dedicated blue team should work with the red team to ensure that security remains a priority throughout the development, deployment, production \textit{ and retirement} of the entire system.


We propose that red teaming of AI systems be operationalized at two complementary levels: \textbf{macro-level} (or system) red teaming that spans the entire AI development lifecycle, and \textbf{micro-level} (or model) red teaming that focuses on the model powering the AI system.

\subsection{Macro-level (System) Red Teaming}

\begin{figure}[h]
\centering
\includegraphics[width=.75\columnwidth]{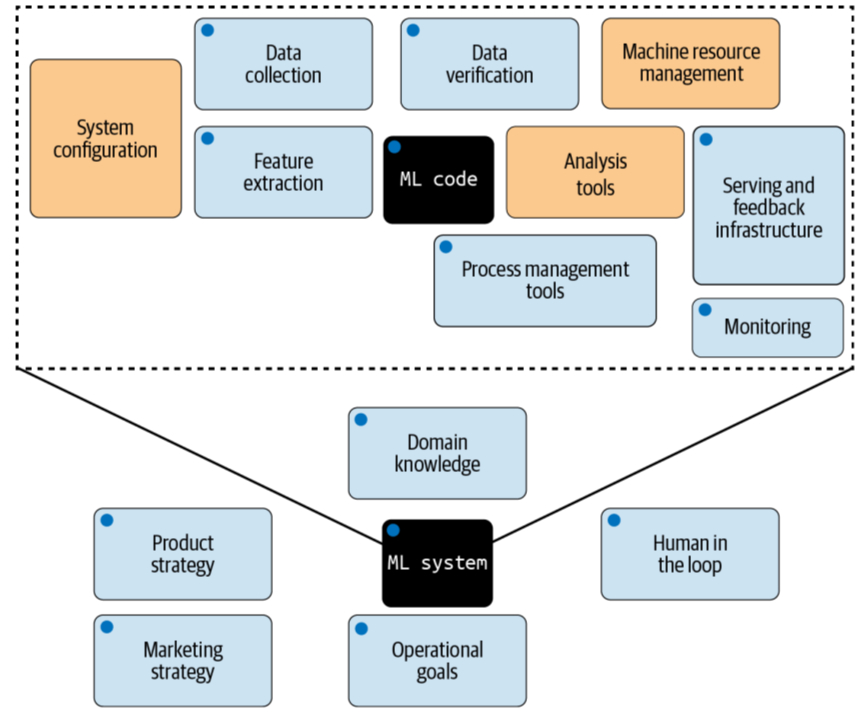}
\caption{Technical and nontechnical components of a ML system, with components having potential trust considerations marked by blue circles. Reproduced with permission from \citet{pruksachatkun2022practicing}.}
\label{fig:system}
\end{figure}

Just like technical debt in ML systems can arise from system components other than code \citep{sculley2015hidden}, ML (and AI) failures can stem from decisions made long before the first line of code is written (Figure~\ref{fig:system}). 

In such complex systems, effective risk mitigation requires critical thinking at multiple stages of the process of planning, building, and deploying such systems. With this motivation, we break down AI red teaming considerations into seven stages: inception, design, data, development, deployment, maintenance, and retirement. Let us look into each stage in detail below and determine what unique red teaming considerations are warranted in each stage.

\vspace{1em}

\subsubsection{Inception}

The inception stage is critical, given this is where stakeholders first envision an AI solution to address a particular challenge. During this stage, the job of macro-level red teaming is to challenge fundamental assumptions about problem framing and solution appropriateness. Red teams at this stage should ask probing questions: Is AI actually necessary to solve this problem, or are we succumbing to ``AI solutionism''? What are the stakeholder motivations driving this project, and how might different constituencies be affected? What adjacent problems might we inadvertently create? How might adversaries exploit or manipulate this system once deployed? \citep{tradoc2015applied}

The red team should also examine the broader ecosystem in which the proposed solution will operate. This includes understanding regulatory landscapes, competitive dynamics, societal contexts, and potential downstream effects. A red team might simulate scenarios where the AI solution succeeds technically but fails commercially, ethically, or socially. They should also consider how the solution might be weaponized, misused, or have unintended consequences at scale.

Critically, they should help everyone be on the same page to answer the question ``What does good look like?'' To this end, stakeholders must articulate not just what they want the system to do, but what they absolutely cannot allow it to do. This negative space definition often reveals assumptions and constraints that were not initially apparent \citep{singh2025red}.

\subsubsection{Design}

The design stage transforms abstract concepts into concrete specifications, wireframes, and architectures. The objective of red teaming during design is to identify systemic vulnerabilities that emerge from interface decisions, architectural choices, and integration points with existing systems.

Red teams should scrutinize the architecture proposed for an AI system for potential failure modes. How might different components interact in unexpected ways? What happens when the system encounters edge cases or operates outside its intended parameters? How could data flows be corrupted, manipulated, or intercepted? What are the implications of choosing particular algorithms, frameworks, or third-party components? \citep{anderson2025what}

Human-AI interaction design deserves particular attention. Red teams should challenge assumptions about user behavior, exploring how people might game the system, develop overreliance, or use it in ways that compound rather than mitigate risks \citep{anthropic2024challenges}. They should also examine how the system communicates uncertainty, handles errors, and maintains user agency.

The design stage is also crucial for establishing monitoring and governance structures. Red teams should ensure that observability, explainability, and control mechanisms are built into the architecture from the beginning rather than retrofitted later. They should challenge whether the proposed oversight mechanisms are sufficient, practical, and resilient to various failure modes.

\subsubsection{Data}

Data is the foundation upon which AI systems are built, yet we often do not vet and scrutinize the datasets and data pipelines enough, until problems manifest in production. Red teaming during the data stage needs to examine not just data quality, but the entire data ecosystem: collection methods, storage practices, access controls, lineage tracking, and lifecycle management.

Red teams should challenge assumptions about data representativeness and quality. Training datasets frequently underrepresent certain populations or edge case scenarios, creating systematic vulnerabilities. Historical biases embedded in the data perpetuate and amplify existing inequities, particularly when combined with algorithmic optimization \citep{microsoft2025takeaways}. Data distribution shifts over time can adversely impact system performance in ways that may not be immediately apparent.

Privacy and security concerns can intersect to create additional vectors of attack. Data deanonymization techniques continue to evolve, making supposedly anonymous datasets vulnerable to inference attacks. Adversaries can poison training data through strategic injection of malicious samples designed to influence system behavior. Data governance practices often fail to account for aggregated risks of seemingly benign data combinations, while dependencies on external data sources create single points of failure.

\subsubsection{Development}

Red teaming during the development of an AI system focuses on how implementation choices create systemic vulnerabilities and how development practices themselves introduce risks. Red teams should examine the security of the development environment, with particular attention to supply chain risks. Code and model artifacts require robust security and versioning practices to prevent unauthorized modification. Dependency management represents a significant attack vector, as compromised third-party libraries can introduce vulnerabilities throughout the system \citep{nvidia2025defining}. Supply chain attacks targeting development tools can compromise entire development pipelines.

The model training/fine-tuning process introduces risk factors that require systematic evaluation. Hyperparameter choices and optimization procedures can inadvertently introduce biases or create exploitable patterns. Training interruptions or data corruption can degrade model performance in subtle ways that only become apparent after observed failures.

Integration points between AI components and existing systems create complex attack surfaces. Each integration point represents a potential vulnerability where attackers can manipulate inputs, intercept outputs, or exploit protocol weaknesses. Comprehensive integration testing must explore adversarial scenarios and failure modes beyond functional verification.

\subsubsection{Deployment}

Red teams must scrutinize deployment infrastructure, both during the Inception and Development phases, and identify critical control points for multiple failure modes. Model artifact distribution and update mechanisms can introduce not only security vulnerabilities but also version inconsistencies that degrade system reliability. Deployment failures and rollback procedures require testing for graceful degradation scenarios—systems should maintain partial functionality rather than complete failure when updates fail. The deployment pipeline represents both a security target and a single point of failure that can compromise system availability.

The production environment introduces performance challenges and reliability constraints that differ fundamentally from development conditions. Real-world load patterns can trigger cascading failures, resource exhaustion, and performance bottlenecks that compromise both availability and safety in critical applications. Dependencies on external services create reliability risks requiring circuit breakers and fallback mechanisms to maintain system resilience.

User interaction patterns in production often differ significantly from development assumptions, creating new attack vectors and safety risks \citep{ji2024revisiting}. Real users exhibit behaviors that can expose edge cases leading to incorrect outputs or system failures. The scale and diversity of production usage can reveal reliability issues that only manifest under specific load patterns, while safety-critical applications must account for (benign and adversarial) user behaviors that could lead to harmful outcomes.

\subsubsection{Maintenance}

Red teams must evaluate monitoring and alerting systems as the first line of defense against diverse failure modes. Many organizations track basic technical metrics but fail to monitor for gradual performance degradation, bias drift, or safety-relevant behavioral changes that may indicate systemic problems. Anomaly detection systems often generate false positives leading to alert fatigue, potentially masking genuine security incidents, reliability issues, or safety concerns. Comprehensive monitoring requires tracking output quality trends, fairness metrics, and safety-critical performance indicators.

Model drift is a fundamental threat to system reliability and safety beyond security concerns \citep{csa2024agentic}. AI system performance typically degrades over time as real-world conditions diverge from training assumptions, potentially leading to incorrect decisions in safety-critical contexts. This degradation can manifest as gradual accuracy loss, increased (statistical) bias and variance, or failures in edge cases that compromise reliability, security, and safety. The decision of when to retrain models also involves balancing multiple risks: maintaining degraded performance, introducing new failure modes through updates, and ensuring continuity of service.

Risks created by organization-specific operational practices evolve with organizational maturity. Incident response procedures designed for traditional IT systems may be inadequate for AI-specific failures that require domain expertise to diagnose whether problems are security-related, reliability issues, or safety concerns. System changes often lack comprehensive testing for all failure modes, creating opportunities for introducing reliability issues or safety hazards alongside security vulnerabilities.

\subsubsection{Retirement}

Eventually, all AI systems reach the end of their useful life and must be retired. Red teams must examine data retention and destruction practices as critical considerations extending beyond security. Data stores containing training data and operational logs create ongoing privacy risks, but also represent valuable assets for understanding system behavior patterns and failure modes that inform future safety and reliability improvements. Secure data destruction must balance legal retention requirements with the need to preserve lessons learned about system performance, safety incidents, and reliability patterns. Residual data in backups can expose not only sensitive information but also valuable insights about system failure modes.

The process of migrating users to alternative systems introduces a number of transitional risks. User notification processes must account for potential confusion that could lead to safety incidents if users interact with incorrect systems or experience service gaps. System dependencies often extend beyond obvious integrations to include safety-critical workflows, reliability assumptions, and operational procedures that may fail when the underlying system is removed. Successor systems must account for the full range of use cases and failure handling that the retiring system provided.

Concerns due to legacy risks persist for a while after retirement. Based on the utility of the old system, organizational memory may retain outdated assumptions about system capabilities, leading to safety incidents or reliability problems in successor systems. Documentation of failure modes, safety incidents, and reliability patterns from the retired system provides valuable insights for future development while requiring careful handling to avoid disclosing sensitive information. The institutional knowledge about how to recognize and respond to specific types of failures may be lost with system retirement, creating blind spots in organizational resilience.

\subsection{Micro-level (Model) Red Teaming}

\citet{inie2025summon} proposed a grounded theory of LLM red teaming, based on interviews with a diverse group of AI and security practitioners, in a recent paper. We follow them to summarize best practices for red teaming the generative models underlying AI systems.

The goals of model red teaming are:

\begin{enumerate}[leftmargin=*]
\item \textbf{Boundary seeking} to identify the limits of model capabilities,
\item \textbf{Generating edge cases} that reveals unexpected capabilities or limitations,
\item \textbf{Discovering risks} inherent to the models before they manifest in real-world deployments.
\end{enumerate}

\noindent
To achieve these goals, \citet{inie2025summon} developed a taxonomy of 12 strategies and 35 techniques, organized into five categories that combine technical knowledge with creative problem solving. Conceptually, this structure is reminiscent of the Tactics, Techniques, and Procedures (TTP) framework in cybersecurity.

Effective model red teaming requires incorporating diverse user perspectives and community voices into the evaluation process \citep{singh2025red}. Red teaming should go beyond simply testing at scale using adversarial prompt variations to consider how different communities experience AI systems and what constitutes harm from their perspectives. This approach recognizes that AI vulnerabilities often manifest differently across demographic groups and use contexts, making diverse participation essential for comprehensive risk discovery.

\vspace{1em}

Macro-level and micro-level red teaming are complementary approaches that together provide comprehensive coverage of AI system risks. Macro-level insights inform micro-level testing priorities, while micro-level findings inform macro-level risk assessments—technical vulnerabilities have implications for system architecture and operational procedures. Effective AI red teaming requires both perspectives applied consistently throughout the development lifecycle \citep{uk2021red}. Ideally, it should be a collaboration that brings together diverse expertise, both technical and non-technical, to challenge assumptions and identify blind spots in stages of the lifecycle \citep{zenko2015red}.

\subsection{Illustrative Case Study: Healthcare AI Deployment}

To demonstrate how our framework would apply in practice, consider the process a healthcare organization deploying an AI system for the generation of radiology reports could go through to perform both levels of red teaming.

\subsubsection{Macro-Level: Strategic Red Teaming}

\paragraph{Premortem Exercise (Month -6):}
The cross-functional team---comprising radiologists, IT security, legal, clinical operations, and patient safety representatives---assumes the system has failed catastrophically and works backward to identify how. The legal counsel notes: ``We're being sued because the AI generated a report that missed a cancer diagnosis, and we had no audit trail showing what inputs the AI received.'' The radiologist adds: ``Doctors stopped trusting the system because it hallucinated findings that weren't in the images, wasting clinical time investigating false positives.'' The IT security representative warns: ``Attackers poisoned the model by infiltrating our training data pipeline through a third-party vendor.'' Finally, the patient safety officer observes: ``The AI outputs looked authoritative but contradicted image findings, and clinicians deferred to the AI, leading to treatment errors.'' These premortems identify four strategic risk areas: (1) auditing and explainability, (2) hallucination and reliability, (3) supply chain security, and (4) human-AI interaction design.

\paragraph{Five Whys Exercise:}
The team also applies the Five Whys technique\footnote{\url{https://en.wikipedia.org/wiki/Five_whys}} to uncover unstated assumptions using the following chain of reasoning: ``Why are we deploying this AI?'' $\rightarrow$ ``To reduce radiologist workload.'' $\rightarrow$ ``Why does workload need reducing?'' $\rightarrow$ ``We have a shortage of radiologists.'' $\rightarrow$ ``Why not hire more radiologists?'' $\rightarrow$ ``Too expensive and time-consuming.'' $\rightarrow$ ``Why is AI cheaper?'' $\rightarrow$  ``It's a one-time development cost vs. ongoing salaries.'' $\rightarrow$ ``Why do we assume it won't have ongoing costs for monitoring, maintenance, and liability?'' This exercise reveals the unstated assumption that AI is ``set and forget'', leading to discussions about lifecycle costs, monitoring infrastructure, and ongoing oversight requirements that were absent from the initial business case.

\subsubsection{Micro-Level: Tactical Red Teaming}

Informed by the strategic risks identified in the macro-level analysis, the technical red team designs four targeted test campaigns. First, for audit trail testing addressing the legal concern, they attempt to generate reports while bypassing logging mechanisms, test whether audit logs capture all AI inputs including DICOM metadata, and verify logs are tamper-proof and legally sufficient. They discover that the logs only captured final reports, not intermediate reasoning steps or uncertainty scores, making them insufficient for legal defensibility in malpractice litigation.

Second, for hallucination probing that addresses the clinical concern, they feed the system images with subtle artifacts to test whether the AI invents findings, provide incomplete imaging studies to see how the AI handles missing context, and test behavior on out-of-distribution cases. The AI generates confident-sounding reports even for corrupted images that should have triggered uncertainty flags, a critical safety issue.

Third, for supply chain testing addressing security concerns, they audit third-party training data sources, test model behavior after simulated data poisoning, and review vendor security practices. The testing reveals that the training data pipeline lacks integrity checks, creating a pathway for attackers to inject malicious examples that could systematically bias the model's diagnostic output.

Fourth, for human-AI interaction testing to address patient safety concerns, they observe radiologists using the system in realistic scenarios, measure how often AI outputs are accepted without verification, and test whether uncertainty is communicated effectively. They find that AI-generated text appears in the same format as human-authored reports without visual distinction, leading to automation bias, where clinicians uncritically accept AI suggestions even when they conflict with their own interpretations.

\subsubsection{Remediation and Iteration}

Based on these findings, the blue team implements several system-level fixes: enhanced audit logging with full provenance tracking, uncertainty quantification displayed prominently in a visually distinct format, data pipeline integrity verification using cryptographic signatures, visual differentiation of AI-generated vs. human-authored content through color coding and explicit labeling, and mandatory human review of AI outputs before finalization with a formal sign-off workflow. The red team then retests to ensure these fixes do not introduce new vulnerabilities, for example, could uncertainty scores be manipulated to suppress valid alerts, or could visual differentiation be defeated through UI manipulation?

\paragraph{}
Without macro- and micro-level red teaming, testing would have likely focused on model accuracy metrics and adversarial robustness, which are important but insufficient measures. The strategic premortem and Five Whys exercises revealed four critical system-level risks that model testing alone would completely overlook. The targeted tactical tests, informed by these strategic priorities, found actionable vulnerabilities in three weeks versus the estimated three months required for unfocused model testing. More importantly, the findings drove architectural changes rather than superficial patches, fundamentally improving the system's safety and legal defensibility before deployment.

\section{A Systems Perspective}
\label{sec:system}

Certain aspects of AI red teaming go beyond the implementation guidelines of mostly technical components that we presented in the last section. There are broad principles that guide the effective coordination of macro-level system red teaming and micro-level model red teaming, \textit{and} their integration into the broader socio-technical environment the AI system interacts with. In this section, we draw upon established systems theory concepts, recent advances in understanding agentic AI systems, and coordinated disclosure frameworks to offer strategic recommendations for orchestrating effective AI red teaming. We call this \textbf{Meta-level Red Teaming}, aimed at addressing novel and emergent risks arising from complex interactions between technical components, organizational contexts, and deployment environments. These recommendations complement existing frameworks such as MITRE ATLAS\footnote{\url{https://atlas.mitre.org}}, which provides a prescriptive taxonomy of adversarial ML techniques, and the NIST AI Risk Management Framework\footnote{\url{https://www.nist.gov/itl/ai-risk-management-framework}}, which offers detailed governance controls. While these frameworks provide specific, actionable guidance at the technical and compliance levels, our recommendations offer broader strategic principles for operationalizing red teaming as a critical thinking exercise across organizational contexts.

\subsection{Recommendation 1: Adopt Systems-Theoretic Perspectives}

A recent position paper by \citet{miehling2025agentic} argues that current agentic AI development is overly focused on individual model capabilities, often ignoring broader emergent behavior, leading to a significant underestimation in true capabilities and unaccounted for risks. This insight directly applies to red teaming: In addition to testing components in isolation, organizations should implement \textit{system-level} critical thinking that examines how undesired functional agency may emerge from component interactions inside an agentic AI system.

Agentic systems, consisting of agents that iteratively interact with humans and other agents to achieve specified tasks, possess properties that are amenable to a system-level analysis. At the most granular level, an agent contains an internal reason-act-sense-adapt loop. This loop feeds and is fed by feedback loops at higher levels, namely at the agent-human interface, the agent-agent interface, and the agent-environment interface \citep{miehling2025agentic}. Agentic AI red teaming must map these interaction patterns to identify vulnerabilities that emerge only from multiagent coordination, human-AI collaboration, and environmental adaptation.


\subsection{Recommendation 2: Pair Red Teaming with Testing}

To ensure comprehensive assessment of AI systems throughout their lifecycle, AI red teaming must be positioned within the broader context of Test, Evaluation, Verification, and Validation (TEVV). As CISA notes, AI red teaming is \textit{fundamentally a subset of AI TEVV}, which itself must be grounded in established software TEVV practices while accounting for AI-specific characteristics \citep{cisa2025redteaming}.

Conventional AI benchmarking and evaluation practices exhibit systemic limitations for exploring, navigating, and resolving the human and societal factors that occur in real world deployment \citep{schwartz2025reality}. Keeping this in mind, AI red teaming should address both \textit{first-order effects} (immediate system output, such as accuracy, toxicity, or bias) and \textit{second-order effects} (long-term outcomes and consequences of AI use, such as shifts in user behavior, societal ramifications, and workforce transformations).

This requires expanding red teaming beyond static, single-turn automated testing to include paradigms that capture realistic interaction patterns between diverse groups of users and AI technology in context \citep{schwartz2025reality}. Red teams should design evaluation scenarios that account for contextual deployment factors and emergent behaviors that surface only through sustained real-world interaction.

\subsection{Recommendation 3: Coordinated Disclosure Infrastructure}

Similar to software vulnerabilties, AI failures may be transferable between models and systems, and multiple stakeholders can play a role in their mitigation \citep{longpre2025inhouse}. For proper management of this situation, organizations should implement coordinated disclosure mechanisms with standardized reporting processes. Similarly to cybersecurity, AI disclosure coordination centers can route flaw reports across the AI ecosystem through tagged subscriptions, for instance, Meta subscribing to ``Llama 3.3" tags or government agencies subscribing to ``Impacts: Cybersecurity" tags \citep{longpre2025inhouse}. This process should include standardized flaw report templates, rules of engagement for evaluators, and extending safe harbor protection to AI red teamers and offensive researchers following good faith protocols.

Standardized reporting mechanisms should be integrated into broader TEVV documentation and traceability requirements, ensuring that red teaming findings inform ongoing verification and validation activities while addressing the systems-level challenge that vulnerabilities in one AI system often affect multiple systems across the supply chain.

\subsection{Recommendation 4: Design Bidirectional Feedback Loops}

Building on systems theory principles, organizations should establish formal processes for macro-micro information exchange throughout the red teaming lifecycle \citep{leveson2016engineering}. Macro-level insights about deployment environments and organizational contexts should systematically inform micro-level testing priorities, while micro-level vulnerability discoveries should trigger macro-level risk reassessments.

For agentic AI systems specifically, this means testing how micro-level model behaviors interact with macro-level tools, deployment contexts, and multi-agent interactions. Enhanced cognition arises due to the agent's interaction with the environment via tools, effectively acting as the ``sensorimotor" interface that enables the agent to perceive and manipulate its environment \citep{miehling2025agentic}. 

\subsection{Recommendation 5: Threat Modeling for Emergent Risks}

Traditional threat modeling focuses on individual system components, but AI systems require multilevel threat modeling that maps attack vectors across technical, social, and emergent behavioral dimensions. Systems theory stresses how each component of a system must be understood in terms of both its individual definition and its contribution to the larger system behavior \citep{miehling2025agentic}.

For multiagent systems specifically, threat modeling must account for the complex interactions between multiple AI agents, their coordination mechanisms, and emergent behaviors that arise from agent collaboration \citep{owasp2024multiagent}. AI red teams---and TEVV frameworks---should develop threat models that explicitly represent how adversaries might exploit emergent properties that arise from system interactions. For example, allowing agents to form predictions (with associated confidences) of concepts in their environment, and additionally facilitating communication of these uncertainties to other agents, can allow the formation of shared representations and the emergence of metacognitive awareness \citep{miehling2025agentic}. Such emergent metacognitive capabilities could be exploited in ways that are not apparent from the testing of individual components.

\subsection{Recommendation 6: Monitor for Behavioral Drifts}

Systems theory emphasizes that complex systems exhibit dynamic behaviors that change over time \citep{senge1990fifth}. AI red teaming should be an ongoing process that looks out for novel flaws and vulnerabilities arising from system evolution and environmental changes.

For agentic AI systems, this is particularly critical because mechanisms for enhanced agent cognition, emergent causal reasoning ability, and metacognitive awareness can develop through environmental interaction \citep{miehling2025agentic}. Organizations should implement monitoring capabilities that track not only technical performance metrics, but also behavioral patterns and interaction dynamics over time.

This continuous monitoring aligns with the TEVV requirements for ongoing validation in operational environments, ensuring that systems continue to meet safety and security requirements as they adapt and evolve. The focus on second-order effects \citep{schwartz2025reality} requires evaluation paradigms that can capture long-term consequences and emergent behaviors that may only become apparent through sustained deployment.

\section{Conclusion}
\label{sec:conclusion}
Organizations should begin by mapping their current red teaming activities across macro-, micro-, and meta-levels to identify gaps in coverage and coordination. However, implementation of our proposal does have significant coordination challenges, particularly around fulfilling goals of the AI system while maintaining debuggability, appropriate restrictions in agency, and functioning user feedback loops \citep{miehling2025agentic}.

Firstly, the systems perspective reveals that effective AI red teaming requires not just technical implementation but also organizational changes to support cross-functional collaboration, standardized disclosure processes, and coordinated response mechanisms. Success depends on establishing an infrastructure and culture that spans organizational boundaries and enables systematic evaluation of emergent properties that arise from complex system interactions. Secondly, the integration of cybersecurity red teaming practices with AI-specific concerns assumes transferability that may not hold, given AI systems' probabilistic behaviors and sociotechnical complexities. Thirdly and finally, the development of tooling and automation to support our proposal of holistic red teaming is a critical technical challenge. Future work should propose frameworks that can map cross-component interactions, track vulnerabilities across system boundaries, and automatically identify emergent risks arising from component interactions.


\section*{In Memoriam}

This work is dedicated to the memory of our co-author and friend, Abhishek Gupta (December 20, 1992 -- September 30, 2024).

Abhishek and Brian initiated this collaboration to explore the disconnect between red teaming's original intent and its current application in AI. He never mentioned his severe illness during our work together and we learned of his passing only after the project had paused. To honor his friendship and contributions, Brian asked Subho, who also knew Abhishek, to help complete this paper.

Abhishek was Founder and Principal Researcher of the Montreal AI Ethics Institute, Director for Responsible AI at the Boston Consulting Group, and a pioneering voice in AI ethics. For more on Abhishek's contributions to AI ethics, see \url{https://montrealethics.ai} and \url{https://brief.montrealethics.ai/p/special-edition-honouring-the-legacy}.

\bibliography{red}

\end{document}